\documentclass{article} 
\usepackage{iclr2020_conference,times, graphicx}

\usepackage{amsmath,amsfonts,bm}

\def\eqref#1{equation~\ref{#1}}

\def\1{\bm{1}}

\DeclareMathAlphabet{\mathsfit}{\encodingdefault}{\sfdefault}{m}{sl}
\SetMathAlphabet{\mathsfit}{bold}{\encodingdefault}{\sfdefault}{bx}{n}

\usepackage{hyperref}
\usepackage{url}
\usepackage{subfig} 

\title{Apple defect detection using Deep Learning-based Object Detection for better post harvest handling}

\author{Paolo F. Valdez \\
Electrical and Electronics Engineering Institute\\
University of the Philippines Diliman\\
Quezon City, NCR, Philippines \\
\texttt{\ paolo.valdez@eee.upd.edu.ph}}

\iclrfinalcopy 
\begin{document}

\maketitle

\begin{abstract}

The inclusion of Computer Vision and Deep Learning technologies in Agriculture aims to increase the harvest quality, and productivity of farmers. During postharvest, the export market and quality evaluation are affected by assorting of fruits and vegetables. In particular, apples are susceptible to a wide range of defects that can occur during harvesting or/and during the post-harvesting period. This paper aims to help farmers with post-harvest handling by exploring if recent computer vision and deep learning methods such as the YOLOv3 (\citet{YOLOv3}) can help in detecting healthy apples from apples with defects.

\end{abstract}

\section{Introduction}

Maintaining high-quality crops is of great importance for the apple industry due to ever-increasing consumer awareness of food quality. Quality inspection is needed due to increasing demand since apples are one of the most consumed fruits globally (\cite{Harker2003}). Post harvest sorting of apples is a difficult, labor intensive process in the industry. The uniformity in size, shape and other quality parameters of apples are required for deciding the overall acceptance quality for customers (\cite{Nissen2016}). Currently many industries perform grading of apples manually. Labor shortages and a lack of overall consistency to the process resulted in a search for automated solutions. The use of computer vision has gained interest in the industry as a fast, reliable, and labor-inexpensive solution to the given problem.

\section{Literature Review}

For more than three decades, apple defect detection has been an interesting area of research. Defect detection is still a challenging task due to huge variation of defect types Computer vision is considered as a useful and practical tool for apple defect detection because of its simplicity, consistency, low cost and high speed (\cite{yang1996}). Despite the limitations of computer vision, it is worth noting that consumers still judge the commercial value of an apple based on the absence of any external defects. 

Classical approaches to computer vision for agriculture uses machine learning. A good example is this segmentation algorithm oriented for regions in citrus fruits (stem, peel, and defects) for detecting the most common peel defects (\cite{blasco2007}). This algorithm that focuses on the difference in contrast on different regions of the citrus fruit, rather than individual pixels was able to achieve a detection accuracy of 95.0 \%.

Different machine learning models such as Support Vector Machine (SVM), Multi-Layer Perceptron (MLP) and K-Nearest Neighbor (KNN) classifiers were comapred in the grading of apples (\citet{moallem}). Two classification tasks are performed: (1) an input apple image is classified as healthy or defected, and (2) an input apple image is classified as first rank, second rank or rejected.  In the two classification tasks, SVM performed best at 92.5\% and 89.2\% accuracies respectively. In another study, A SVM model based on particle swarm optimization has also attempted apple grading and have reported maximum accuracy rate of 91\% (\cite{Ji2018}).

Modern approaches employ deep-learning based approach.
Recently, Tian et al. have proposed a combination of YOLOv3 and Cycle-Consistent Adversarial Network (CycleGAN) was used in detecting of apple lesions \citet{Tian2019}. Their work approached it as an image classifaction problem wherein their model is trained
on a dataset of 640 images composed of healthy apples and apples with defects. The images collected in two ways: orchard field collection and online collection. Data augmentation techniques like Cycle-Consistent Adversarial Network (CycleGAN) is used to artificially
expand the dataset. DenseNet is used as a feature extractor
to enhance the detection results of the YOLO-v3 model.
This is the first and also the most recent study that has
proposed a member of the YOLO family algorithm for apple defect detection.

Previous studies view fruit defects as an image classification problem rather than an object detection problem. This paper aims to approach apple defects as an object detection problem by using a data set that contains healthy apples and apples with defects to train a YOLO-v3 model in order to detect which apples are healthy. The goal of which to improve post-harvest handling and help farmers improve their productivity  

\section{YOLOv3 Architecture}

The deep learning algorithm used belongs to the“You Only Look Once (YOLO)” family of algorithms. This family of algorithms leverages the use of convolutional neural networks for object detection. They are one of the faster object detection algorithms available and a good choice for real-time detection, without the trade-off of losing accuracy.

\begin{figure}[!htb]
\begin{center}
\includegraphics[width=0.8\textwidth]{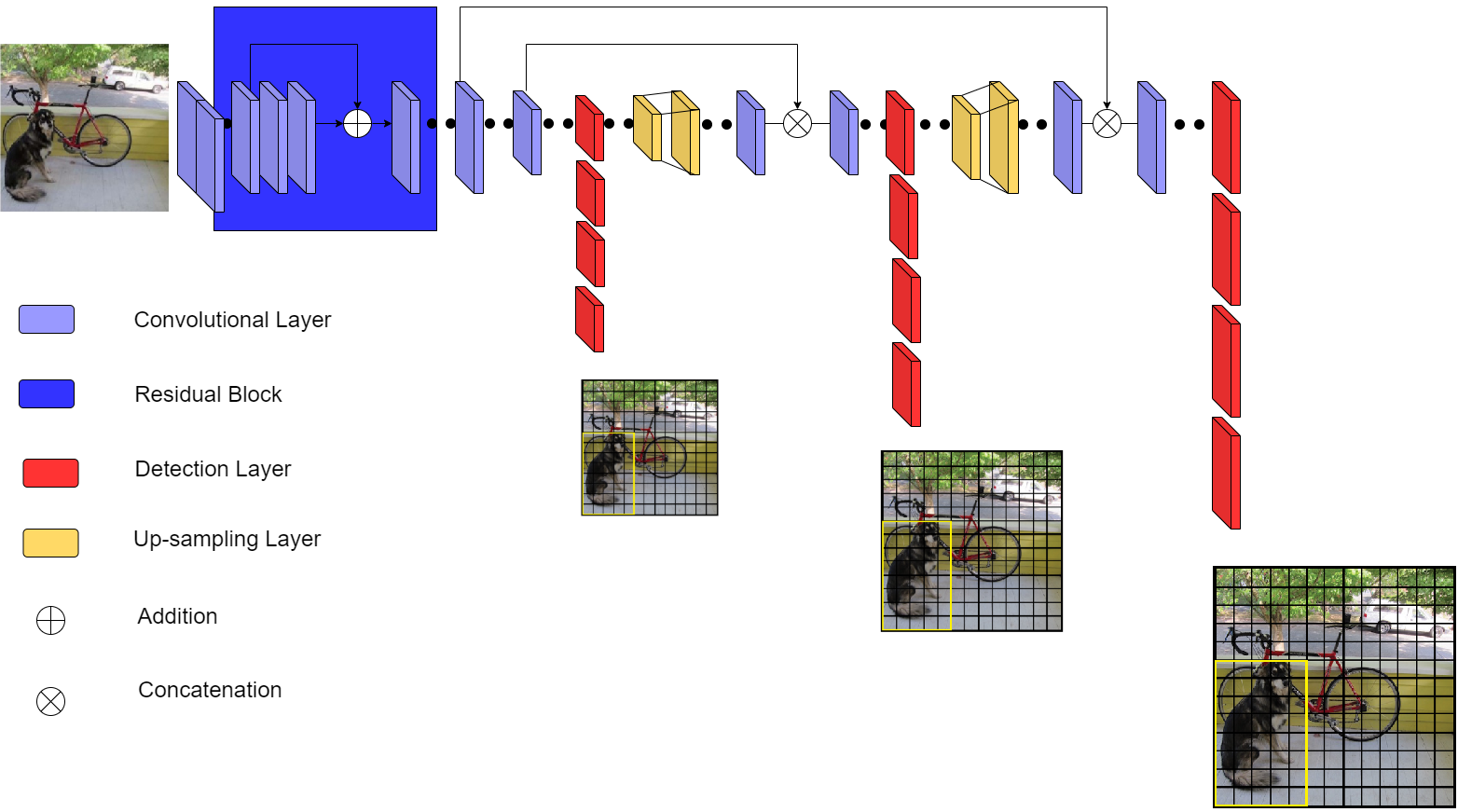}
\end{center}
\caption{The YOLO-v3 architecture}
\end{figure}

The YOLO-v3 architecture is shown in Figure 1. Feature-learning is done through the convolutional layers. No fully-connected layer is used, thus this model is image size agnostic. Also, no pooling layers are used. In order to pass size-invariant features forward, a convolutional layer with a given stride is used to downsample the feature map (\cite{YOLOv3}). The ResNet-like structure in the YOLOv3 are called Residual Blocks and are used for feature learning. The Residual Blocks consists of several convolutional layers and skip connections. The unique feature of YOLOv3 is that it makes detection at three different scales.

\section{SSD: Single Shot Detector}

In the SSD (\cite{SSD}) , as the name suggest, both the object localization and classification task are performed in a single forward pass of the network. It has no delegated region proposal network and predicts both the boundary boxes and the classes directly from feature maps. The SSD predicts the offset distance of predefined anchor boxes for every location of the feature map. The architecture of the SSD is based on the VGG-16 architecture, but discards the fully connected layers.

\section{Dataset}

The images used for this experiment are gathered through online collection. The custom training set is composed of 452 images of apples, 226 images for healthy apples and 226 images for apples with defect. The test set is composed of 140 images composed of 70 images for health apples and 70 images of apples with defect.

\begin{figure}
\centering
\begin{minipage}{.5\textwidth}
  \centering
  \includegraphics[width=.5\linewidth]{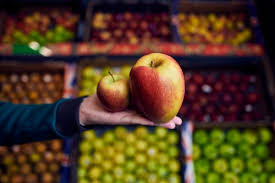}
  \captionof{figure}{Sample image from Healthy Apple}
  \label{fig:healthy-apple}
\end{minipage}%
\begin{minipage}{.5\textwidth}
  \centering
  \includegraphics[width=.4\linewidth]{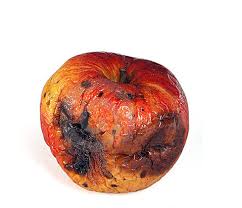}
  \captionof{figure}{Sample image Apple with Defects}
  \label{fig:apple-with-defects}
\end{minipage}
\end{figure}

\subsection{Materials and Methods}

The experiment presented in this paper is carried out using Python version 3.6.7 on a NVIDIA GTX 1060 machine with Intel Core i7-7700HQ processor and 16GB RAM. The computing platform used was the Nvidia CUDA 10.0.0. In training the YOLOv3 model, the images serves as input to a CNN model and the resulting detections are filtered using the non-max suppression algorithm. To speed up training, I used a pre-trainined model on the COCO data set (\cite{COCO2014}) and then trained on the custom data set. For the SSD-based object detector, I trained a model based from the Tensorflow Object Detection API, which is an open source framework. For this model, I also used a pre-trainined model on the COCO data set and then trained on the custom data set.

\section{Metrics Used}

The metrics used for this experiment are the average precision (AP) of the class of apple (healthy or with defect) and mean average precision (mAP). To derive these the following concepts must first be understood.

\begin{itemize}
    \item Intersection Over Union (IOU) is defined as the area of overlap between the predicted bounding box (Bp) and the ground truth bounding box (Bgt) divided by the area of union between them. This is expressed by the formula:
    
    $ IOU= \frac{ B_p \cap B_{gt} }{ B_p \cup B_{gt}}$

	\item True Positive (T\textsubscript{p}) refers to the right detection of our chosen model i.e. a detection where IOU is greater than or equal to IoU threshold.
    \item False Positive (F\textsubscript{p}) refers to the erroneous detections of our chosen model i.e. a
detection where IOU is less than IOU threshold.
    \item False Positive (F\textsubscript{p}) refers to an incorrect detection i.e. a detection where IOU is less than IOU threshold.
    
    \item Precision (P) is defined as the number of true positives (T\textsubscript{p}) divided by the sum of true positives (T\textsubscript{p}) and false positives (F\textsubscript{p}).
    
    $P = \frac{T_p}{T_p + F_p}$

    \item Recall (R) is defined as the number of true positives (T\textsubscript{p}) divided by the sum of true positives (T\textsubscript{p}) and false negatives (F\textsubscript{n}).
    
    $R = \frac{T_p}{T_p + F_n}$
    
    \item Average Precision (AP) is the precision averaged across all recall values ranging from 0 to 1.
    
    \item Mean Average Precision (mAP) is computed by taking the AP for each class (healthy apple, apple with defects) and averaging them.
\end{itemize}

\section{Results}

As shown in table 1, the SSD-based detector has better performance in detecting healthy apples as the intersection over union (IOU) is increased. The difference between the two is more pronounced in detecting apples with defects. As shown in table 2, YOLOv3 does a better job at detecting the apples with defects due mostly to its architecture.


\begin{table}[!htbp]
\caption{Average Precision (AP) on Healthy Apple detection performance}
\label{table-healthyapple}
\begin{center}
\begin{tabular}{lllll}
\multicolumn{1}{c}{\bf PART}  &\multicolumn{1}{c}{\bf AP@0.3IOU}
&\multicolumn{1}{c}{\bf AP@0.5IOU}
&\multicolumn{1}{c}{\bf AP@0.7IOU}
\\ \hline \\
YOLOv3-based &0.8078 &0.8035 &0.7387\\
SSD-based &0.8042 &0.8042 &0.7726\\
\end{tabular}
\end{center}
\end{table}


\begin{table}[!htbp]
\caption{Average Precision (AP) on Apple with Defects detection performance}
\label{table-applewithdefects}
\begin{center}
\begin{tabular}{lllll}
\multicolumn{1}{c}{\bf PART}  &\multicolumn{1}{c}{\bf AP@0.3IOU}
&\multicolumn{1}{c}{\bf AP@0.5IOU}
&\multicolumn{1}{c}{\bf AP@0.7IOU}
\\ \hline \\
YOLOv3-based &0.6909 &0.6824 &0.6084\\
SSD-based &0.5951 &0.5875 &0.5613\\
\end{tabular}
\end{center}
\end{table}


\begin{table}[!htbp]
\caption{Mean Average Precision (mAP) Apple detection performance}
\label{results}
\begin{center}
\begin{tabular}{lllll}
\multicolumn{1}{c}{\bf PART}  &\multicolumn{1}{c}{\bf mAP@0.3IOU}
&\multicolumn{1}{c}{\bf mAP@0.5IOU}
&\multicolumn{1}{c}{\bf mAP@0.7IOU}
\\ \hline \\
YOLOv3-based &0.7469 &0.7430 &0.6736\\
SSD-based &0.6997 &0.6959 &0.6667\\
\end{tabular}
\end{center}
\end{table}

 The SSD-based detector has more false positives on the apples with defects as this model has difficulty in detecting smaller apple lesions. Thus resulting to a lower mean average precision (mAP) compared to the YOLOv3-based detector, as shown in table 3.


\begin{figure}[!htb]
\begin{center}

\framebox[4.0in]{\includegraphics[width=0.5\textwidth]{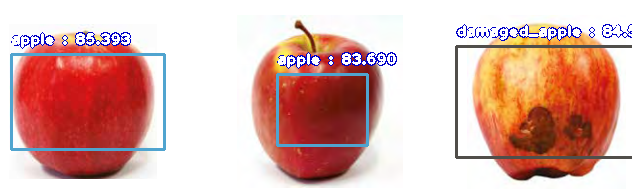}}
\end{center}
\caption{Sample Results from training the YOLOv3-based apple detector}
\end{figure}

The Residual Blocks in architecture of YOLOv3 helps detecting small details such as the apple defects. A sample result is shown in Figure 3. Future works include training it on other types of fruits and crops, using image-to-image translation models as a data augmentation technique in the training set.

\bibliography{iclr2020_conference}

\begin{thebibliography}{10}
\providecommand{\natexlab}[1]{#1}
\providecommand{\url}[1]{\texttt{#1}}
\expandafter\ifx\csname urlstyle\endcsname\relax
  \providecommand{\doi}[1]{doi: #1}\else
  \providecommand{\doi}{doi: \begingroup \urlstyle{rm}\Url}\fi

\bibitem[Blasco et~al.(2007)Blasco, Aleixos, and Molto]{blasco2007}
Jose Blasco, N.~Aleixos, and Enrique Molto.
\newblock Computer vision detection of peel defects in citrus by means of a
  region oriented segmentation algorithm.
\newblock \emph{Journal of Food Engineering}, 81:\penalty0 535--543, 08 2007.
\newblock \doi{10.1016/j.jfoodeng.2006.12.007}.

\bibitem[Harker et~al.(2003)Harker, Gunson, and Jaeger]{Harker2003}
F.Roger Harker, F.Anne Gunson, and Sara~R. Jaeger.
\newblock The case for fruit quality: an interpretive review of consumer
  attitudes, and preferences for apples.
\newblock \emph{Postharvest Biology and Technology}, 28\penalty0 (3):\penalty0
  333 -- 347, 2003.
\newblock \doi{10.1016/S0925-5214(02)00215-6}.

\bibitem[{Ji} et~al.(2018){Ji}, {Zhao}, {Bi}, and {Shen}]{Ji2018}
Y.~{Ji}, Q.~{Zhao}, S.~{Bi}, and T.~{Shen}.
\newblock Apple grading method based on features of color and defect.
\newblock In \emph{2018 37th Chinese Control Conference (CCC)}, pp.\
  5364--5368, July 2018.
\newblock \doi{10.23919/ChiCC.2018.8483825}.

\bibitem[Lin et~al.(2014)Lin, Maire, Belongie, Hays, Perona, Ramanan,
  Doll{\'a}r, and Zitnick]{COCO2014}
Tsung-Yi Lin, Michael Maire, Serge Belongie, James Hays, Pietro Perona, Deva
  Ramanan, Piotr Doll{\'a}r, and C.~Lawrence Zitnick.
\newblock Microsoft coco: Common objects in context.
\newblock In David Fleet, Tomas Pajdla, Bernt Schiele, and Tinne Tuytelaars
  (eds.), \emph{Computer Vision -- ECCV 2014}, pp.\  740--755, Cham, 2014.
  Springer International Publishing.
\newblock ISBN 978-3-319-10602-1.

\bibitem[Liu et~al.(2015)Liu, Anguelov, Erhan, Szegedy, Reed, Fu, and
  Berg]{SSD}
Wei Liu, Dragomir Anguelov, Dumitru Erhan, Christian Szegedy, Scott~E. Reed,
  Cheng{-}Yang Fu, and Alexander~C. Berg.
\newblock {SSD:} single shot multibox detector.
\newblock \emph{CoRR}, abs/1512.02325, 2015.
\newblock URL \url{http://arxiv.org/abs/1512.02325}.

\bibitem[Moallem et~al.(2017)Moallem, Serajoddin, and Pourghassem]{moallem}
Payman Moallem, Alireza Serajoddin, and Hossein Pourghassem.
\newblock Computer vision-based apple grading for golden delicious apples based
  on surface features.
\newblock \emph{Information Processing in Agriculture}, 4\penalty0
  (1):\penalty0 33 -- 40, 2017.
\newblock ISSN 2214-3173.
\newblock \doi{https://doi.org/10.1016/j.inpa.2016.10.003}.
\newblock URL
  \url{http://www.sciencedirect.com/science/article/pii/S2214317315300068}.

\bibitem[Nissen et~al.(2016)Nissen, Bound, Adhikari, and Cover]{Nissen2016}
Robert Nissen, Sally Bound, Rajendra Adhikari, and Ian Cover.
\newblock Factors affecting postharvest management of apples: a guide to
  optimising quality.
\newblock \emph{Small}, 20\penalty0 (53):\penalty0 20, 2016.

\bibitem[Redmon \& Farhadi(2018)Redmon and Farhadi]{YOLOv3}
Joseph Redmon and Ali Farhadi.
\newblock Yolov3: An incremental improvement.
\newblock \emph{CoRR}, abs/1804.02767, 2018.
\newblock URL \url{http://arxiv.org/abs/1804.02767}.

\bibitem[Tian et~al.(2019)Tian, Yang, Wang, Li, and Liang]{Tian2019}
Yunong Tian, Guodong Yang, Zhe Wang, En~Li, and Zize Liang.
\newblock Detection of apple lesions in orchards based on deep learning methods
  of cyclegan and yolov3-dense.
\newblock \emph{J. Sensors}, 2019:\penalty0 7630926:1--7630926:13, 2019.

\bibitem[Yang(1996)]{yang1996}
Qingsheng Yang.
\newblock Apple stem and calyx identification with machine vision.
\newblock \emph{Journal of Agricultural Engineering Research}, 63\penalty0
  (3):\penalty0 229 -- 236, 1996.
\newblock ISSN 0021-8634.
\newblock \doi{https://doi.org/10.1006/jaer.1996.0024}.
\newblock URL
  \url{http://www.sciencedirect.com/science/article/pii/S0021863496900244}.

\end{thebibliography}
\bibliographystyle{iclr2020_conference}

\end{document}